\documentclass{bmvc2k}
\usepackage{parskip}
\usepackage{verbatim}
\usepackage{multirow}
\usepackage{tabularx}

\newcolumntype{L}[1]{>{\raggedright\arraybackslash}p{#1}}
\newcolumntype{C}[1]{>{\centering\arraybackslash}p{#1}}
\newcolumntype{R}[1]{>{\raggedleft\arraybackslash}p{#1}}

\title{VideoNavQA: Bridging the Gap between Visual and Embodied Question Answering}

\addauthor{C\u{a}t\u{a}lina Cangea}{Catalina.Cangea@cst.cam.ac.uk}{1}
\addauthor{Eugene Belilovsky}{eugene.belilovsky@umontreal.ca}{2}
\addauthor{Pietro Li\`{o}}{Pietro.Lio@cst.cam.ac.uk}{1}
\addauthor{Aaron Courville}{aaron.courville@umontreal.ca}{3}

\addinstitution{
 University of Cambridge\\
 Cambridge, United Kingdom
}
\addinstitution{
 Mila, Universit\'{e} de Montr\'{e}al\\
 Montr\'{e}al, Canada
}
\addinstitution{
 Mila, CIFAR Fellow\\
 Montr\'{e}al, Canada
}

\runninghead{Cangea, Belilovsky, Li\`{o}, Courville}{VideoNavQA}

\def\etal{\emph{et al}\bmvaOneDot}

\begin{document}

\maketitle

\begin{abstract}
\noindent{Embodied Question Answering (EQA) is a recently proposed task, where an agent is placed in a rich 3D environment and must act based solely on its egocentric input to answer a given question. The desired outcome is that the agent learns to combine capabilities such as scene understanding, navigation and language understanding in order to perform complex reasoning in the visual world. However, initial advancements combining standard vision and language methods with imitation and reinforcement learning algorithms have shown EQA might be too complex and challenging for these techniques. In order to investigate the feasibility of EQA-type tasks, we build the VideoNavQA dataset that contains pairs of questions and videos generated in the House3D environment. The goal of this dataset is to assess question-answering performance from nearly-ideal navigation paths, while considering a much more complete variety of questions than current instantiations of the EQA task. We investigate several models, adapted from popular VQA methods, on this new benchmark. This establishes an initial understanding of how well VQA-style methods can perform within this novel EQA paradigm.}

\end{abstract}

\section{Introduction}

\begin{figure}[t]
    \centering
    \includegraphics[trim={0, 325, 0, 0}, clip, width=\linewidth]{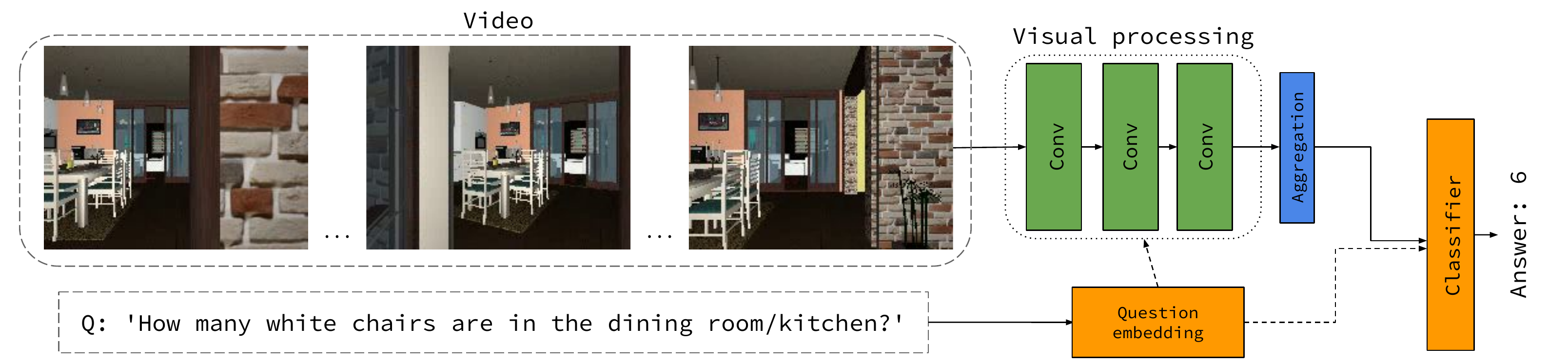}
    \caption{High-level description of the VideoNavQA task and our approach: the VQA system receives an example containing a video of a trajectory inside the house environment and a question, processes the input and produces an answer.}
    \vspace{-0.5em}
    \label{fig:overall}
\end{figure}

The Embodied Question Answering~\cite{embodiedqa} (EQA) and Interactive Question Answering~\cite{gordon2018iqa} (IQA) tasks were introduced as a means to study the capabilities of agents in rich, realistic environments, requiring both \emph{navigation and reasoning} to achieve success. Each of these skills typically needs a different approach~\cite{mishkin2019benchmarking,hu2017learning,hudson2018compositional} that should nevertheless be smoothly integrated with the rest of the system leveraged by the agent. These abilities are assessed via placing the agent at a random location in a house environment and asking it a question---successful completion of the task involves the agent knowledgeably exploring the environment and reasoning about visual stimuli.

Initial attempts at solving the EQA task~\cite{embodiedqa,das2018neural} have thus combined typically used vision (convolutional neural networks for object detection) and language (question encoding or program generation) techniques with imitation learning and reinforcement learning. However, these approaches either suffer from potentially weaker performance than when using a language-only model~\cite{anand2018blindfold} or are preceded by additional hand-engineered steps (manually defined sub-goals, imitation learning on pre-computed expert trajectories). Indeed, even for simple single object questions, the agent is often unable to advance meaningfully towards the target~\cite{das2018neural}, producing visual streams that cannot be used to answer the query. This suggests that EQA is a highly challenging task which might not be easily approached from this angle---further investigation is required to realistically assess the gap between current state-of-the-art and desired human level performance on the EQA performance.

Single-image Visual Question Answering is only now starting to tackle complex reasoning questions, even in limited settings~\cite{hudson2018compositional,johnson2017clevr}---it is unclear if existing methods can handle the rich video streams produced in our task. On the other hand, EQA introduces a navigation component, which can often lead to video inputs which are uninformative with respect to the question. A natural issue arises: can the desired tasks be solved with current methods, if we assume the agent is given correct visual streams (ones that can be used to provide an answer)? In our attempt to answer this question, we propose a dataset that decouples the visual reasoning from the navigation aspects in the EQA problem. We introduce the VideoNavQA task, illustrated in Figure~\ref{fig:overall}. While removing the navigation and action selection requirements from EQA, we increase the difficulty of the visual reasoning component via a much larger question space, tackling the sort of complex reasoning questions that make QA tasks challenging~\cite{johnson2017clevr}. By designing and evaluating several VQA-style models on the dataset, we \emph{establish a novel way of evaluating EQA feasibility} given existing methods, while highlighting the difficulty of the problem even in the most ideal setting.

\section{Related work}

To the best of our knowledge, there are no previous works that position themselves at the intersection of these two paradigms. Instead, VQA and EQA have been tackled from separate angles, with multiple interesting developments in each case.

Das~\etal\cite{embodiedqa} proposed the \textbf{EQA}-v1 dataset containing 4 types of questions (\small \texttt{location}, \texttt{color}, \texttt{color\_room}, \texttt{preposition}\normalsize) that always refer to a single object, the navigation goal of the agent. They train a model using imitation and reinforcement learning, while revealing that RL finetuning often results in overshooting the goal. A subsequent improvement is achieved via hierarchical policy learning with neural modular control~\cite{das2018neural}---however, this approach uses hand-crafted sub-policies. Anand~\etal\cite{anand2018blindfold} study the performance of question-only baselines on EQA-v1, which give better results when the agent is spawned more than 10 steps away, concluding that existing EQA methods struggle to exploit (and are often impeded by) the environment. A more recent extension of EQA, multi-target EQA~\cite{eqa_multitarget}, requires reaching multiple goals to answer the question (for example, comparing the sizes of two objects in different rooms). The authors build the MT-EQA dataset containing 6 types of questions and tackle the task by first decomposing each question into small sub-goals, which are more easily achievable by a model similar to the one used in previous works. Another EQA variant based on photorealistic environments~\cite{eqa_matterport} uses point clouds instead of RGB input; the authors ``port'' three of the EQA-v1 questions to Matterport3D. Although in this work we focus on the House3D environment and EQA, we also note that the IQA task~\cite{gordon2018iqa} for the AI2Thor~\cite{ai2thor} environment was introduced with similar goals. However, this task is defined in single room settings, hence the negative effect of poor navigation is less severe.

\textbf{Visual question answering} has been extensively tackled over the past few years, with numerous datasets being released, various algorithms designed and general studies carried out~\cite{malinowski2018learning, bahdanau2018systematic}. The VQA dataset~\cite{antol2015vqa} is one such example, containing free-form and open-ended questions about real-life images and abstract scenes that require natural language answers. CLEVR~\cite{johnson2017clevr} uses a functional program-based question representation to generate a vast range of questions from synthetic scene graphs that contain 3--10 objects with restricted variability (3 types, 2 sizes, 2 materials, 8 colors). More recently, the GQA dataset~\cite{hudson2019gqa} has been proposed to address some of the issues of current datasets, including biases in the answer distribution. Widely used VQA models include \emph{stacked attention networks}~\cite{yang2016stacked}, which use the question embedding as a query for attending over the visual input, \emph{multimodal compact bilinear pooling}~\cite{fukui2016multimodal}, which fuses the text and visual embeddings via multiplying them in the Fourier space, \emph{feature-wise linear modulation}~\cite{perez2018film}, which selects informative visual feature maps via question-based conditioning, its \emph{multi-hop} extension~\cite{strub2018visual}, which conditions feature maps by iteratively attending over the language input, and \emph{compositional attention networks} (MAC)~\cite{hudson2018compositional}, which can reason about the visual input in a multi-step fashion using the memory-attention-control cell. Graph neural network-based approaches have also been used for VQA tasks lately~\cite{norcliffe2018learning, narasimhan2018out, teney2017graph}, operating on the relational representation of objects in the image. The model that most closely resembles the ones we propose is explored in a multi-turn QA setting~\cite{nguyen2018film}, where the system is provided with a dialogue (set of question-answer pairs) and a video---the question encoding is used to perform both per-frame conditioning and attention over the hidden states of the LSTM which encodes all the video frames.

\begin{table}[t]
\parbox{.3\linewidth}{
\begin{tabular}{|c|C{0.5cm}c|}
\hline
 & \footnotesize Houses & \footnotesize Samples \\
\hline
\footnotesize Train & \footnotesize 622 & \footnotesize 84990 \\
\footnotesize Validation & \footnotesize 65 & \footnotesize 8755 \\
\footnotesize Test & \footnotesize 56 & \footnotesize 7587 \\
\hline
\end{tabular}
\caption{Dataset split statistics. The three sets of environments are disjoint.}
\label{table:splitstats}
}
\parbox{.7\linewidth}{
\begin{center}
\begin{tabular}{C{1.5cm}C{6.55cm}}
  \footnotesize EQAv1       & \footnotesize What room is the <OBJ> located in?\\
  \footnotesize (Q types: 4)       & \footnotesize What color is the <OBJ> in the <ROOM>? \\\hline
         \footnotesize VideoNavQA  & \footnotesize Are both <attr1> <OBJ1> and <attr2> <OBJ2> <color>? \\
        \footnotesize (Q types: 28) & \footnotesize How many <attr> <OBJ> are in the <ROOM>?  \\
         & \footnotesize Is there <art> <attr> <OBJ>?\\
    \end{tabular}
    \caption{Some examples of the type of question templates found in EQAv1 versus the more complex existence, counting and comparison questions found in VideoNavQA.}
    \label{table:qex}
    \end{center}
}
\vspace{-1em}
\end{table}

Several datasets have been proposed that also consider \textbf{video question answering} in settings such as movies~\cite{MovieQA, mun2017marioQA,lei2018tvqa}. They are often provided with rich per-frame annotations (for example, subtitles in movies). However, these tasks largely focus on identifying actions or other dynamic behaviors and are distant from the aims of VideoNavQA. Our task considers indoor navigation trajectories that exhibit rich visual data at each time step. This poses an additional challenge for video QA systems---these are now required to \emph{isolate the relevant information from a large pool of visual concepts} present in each frame and \emph{perform more advanced reasoning to answer the extensive variety of questions} designed.

\section{The VideoNavQA dataset}

\begin{figure}
    \centering
    \includegraphics[trim={0 50 0 0},clip,width=0.34\linewidth]{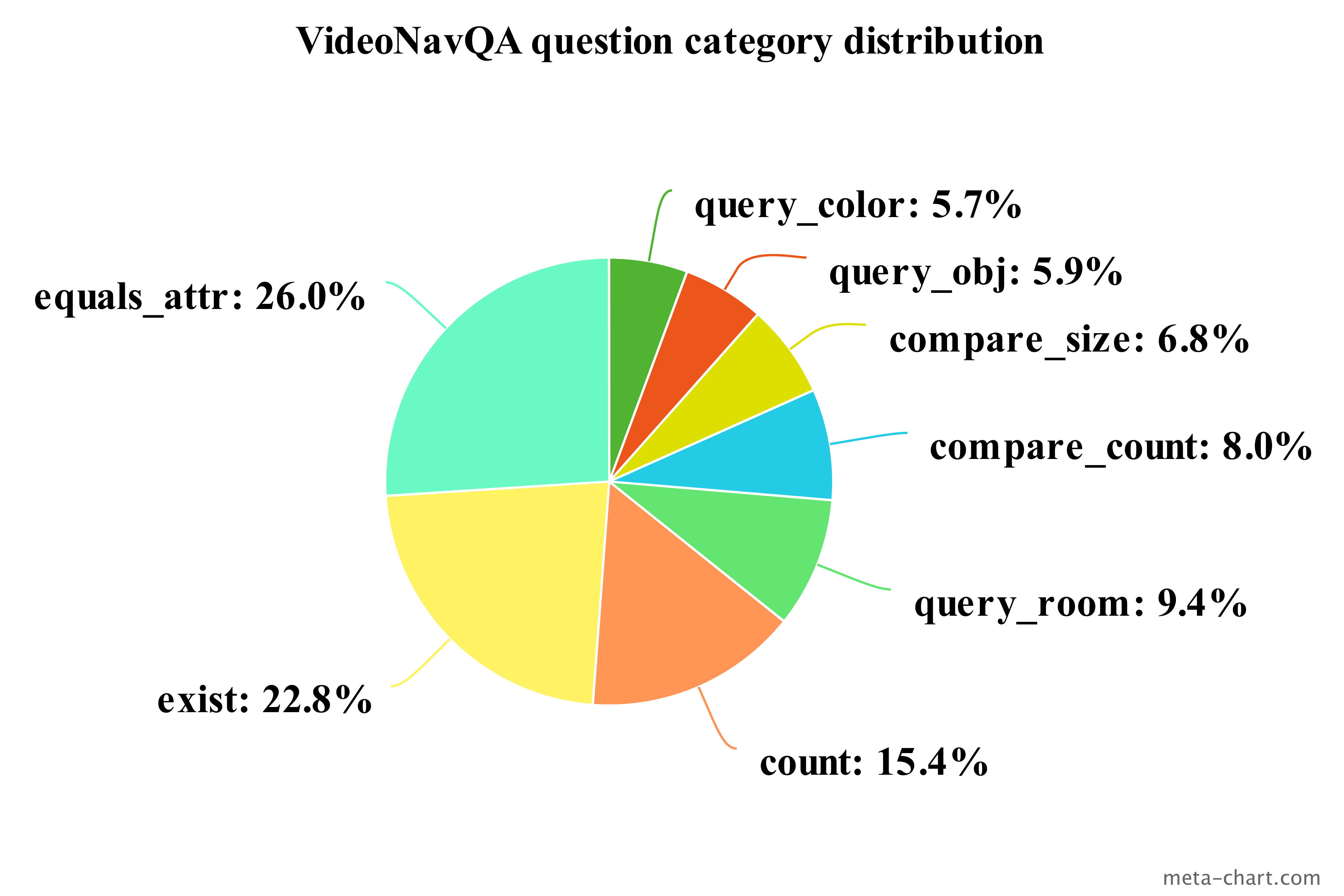}\includegraphics[width=0.33\linewidth]{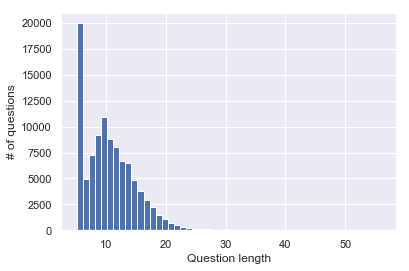}\includegraphics[width=0.33\linewidth]{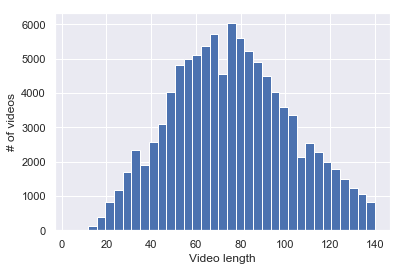}
    \caption{VideoNavQA dataset distribution. (Left:) Proportions for each question category. (Middle:) Question lengths (maximum is 56). (Right:) Video lengths (maximum is 140).}
    \vspace{-1em}
    \label{fig:stats}
\end{figure}

We have constructed a new benchmark to study the capabilities of VQA approaches in a variant of the EQA task, where the agent is required to answer a question while having access to a near-optimal trajectory---that is, the trajectory taken corresponds to natural trajectories with sufficient information in the video input to answer the question. Our task can be seen as complementary to the Habitat Challenge~\cite{savva2019habitat}, where the focus is on navigation instead of question answering. We use the House3D virtual environment~\cite{wu2018building} to generate approximately 101,000 pairs of videos and questions; the dataset contains 28 types of questions belonging to 8 categories (see Figure~\ref{fig:stats}), with 70 possible answers. Each question type is associated with a template that facilitates programmatic generation using ground truth information extracted from the video. The complexity of the questions in the dataset is far beyond that of other similar tasks using this generation method (such as CLEVR): the questions involve single or multiple object/room existence, object/room counting, object color recognition and localization, spatial reasoning, object/room size comparison and equality of object attributes (color, room location). The full list of question types and dataset counts can be found in the supplementary material.\footnote{The dataset and code can be found at \url{https://github.com/catalina17/VideoNavQA}.}

\subsection{Visual information}

\textbf{Environments and scene representation} \enskip The House3D environment is based on indoor scenes from the SUNCG dataset~\cite{song2016ssc}. Table~\ref{table:splitstats} shows the number of houses used in each of the three dataset splits, along with the corresponding number of examples (question-video pairs). Each house appears only in a single split and in no more than 150 videos, to ensure that there is a large and consistent variability in the visual information across the dataset.

\textbf{Video generation and ground truth extraction} \enskip Each example in the VideoNavQA dataset is constructed by first generating the video component. We use the underlying grid representation of the environments to compute the shortest path between arbitrary locations in two different rooms of the house. To obtain the video, we render the shortest-path trajectory inside House3D. This corresponds to what an agent would see in the EQA setting while exploring the house, with the added benefit that its trajectory is already sensible from a navigation perspective. The ground truth information is then obtained by parsing each video frame. We use the SUNCG semantic rendering to identify the objects that are visible. By linking them via depth rendering to the current room the agent is in, or to an adjacent one, we index the objects and rooms that are seen on the trajectory and therefore in the video.

\subsection{Questions}

\textbf{Functional form representation} \enskip Similarly to other synthetic benchmark datasets (EQA-v1~\cite{embodiedqa}, CLEVR~\cite{johnson2017clevr}), we choose to generate questions according to functional, template-style representations (e.g. \emph{``How many <attr> <obj\_type-pl> are in the <room\_type>?''}). This facilitates the instantiation of ground truth tags, once the video for the corresponding trajectory has been generated and analyzed. Moreover, we can easily execute the corresponding program to determine the answer---this amounts to performing a series of basic operations such as \small{\texttt{filter()}, \texttt{count()}, \texttt{get\_attr()}} \normalsize on the ground truth.

\textbf{Generation} \enskip The question generation process starts by randomly choosing one of the 28 templates to be instantiated. A valid question will always have tags instantiated with ground truth values. For example, if there is a \emph{<room\_type>} tag and we have only seen a kitchen and a living room on our trajectory, then the set of possible instantiations is \{\emph{kitchen}, \emph{living room}\}. Using this principle, we build sets of possible values for each tag in the template. In order to generate a valid \emph{(question, answer)} pair, we randomly assign each tag a value from its set, then run the template functional program to compute whether the question is valid and can be answered using the ground truth. To illustrate the process, consider the template \emph{``What color is the <attr> <obj\_type>?''} with the associated program:

\small{\texttt{input(objs)$\rightarrow$filter(obj\_type)$\rightarrow$filter(attr)$\rightarrow$unique()$\rightarrow$get\_attr(color)}}

\normalsize We first select the set of objects seen on the trajectory from our ground truth representation. Next, we filter by the instantiated object type, then by the instantiated attribute (enforced to not be a color during the tag value assignment). Finally, we ensure that the result is unique (i.e. that the question is unambiguous) and retrieve the color of the object as the answer.

\section{Models}

\begin{figure}
    \centering
    \includegraphics[trim={0 300 30 25},clip,width=\linewidth]{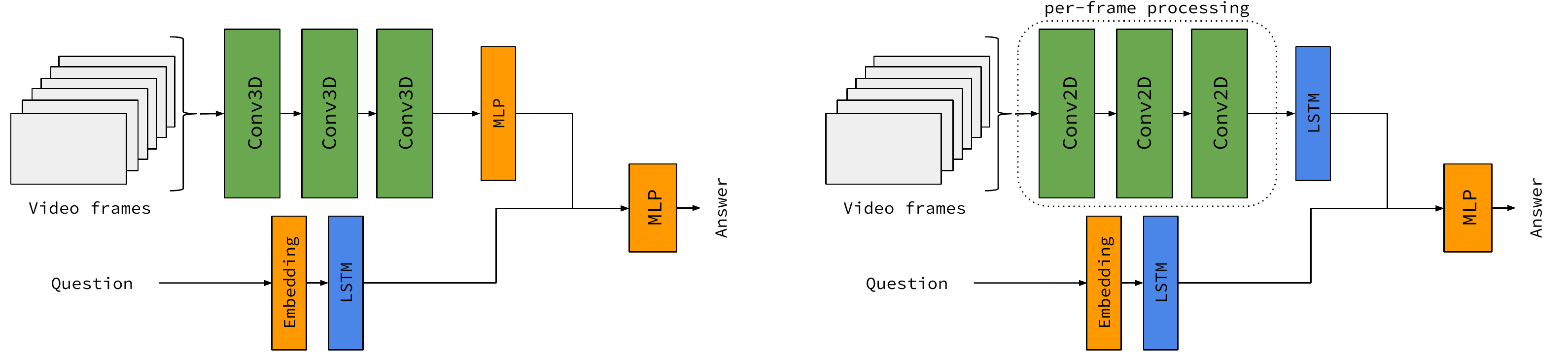}
    \caption{(Left:) Concat-CNN3D processes the entire video. (Right:) Concat-CNN2D aggregates frame features via an LSTM. Both merge the result with the question embedding.}
    \vspace{-1em}
    \label{fig:concats}
\end{figure}

We have designed the VideoNavQA dataset to address the EQA challenge from an alternative perspective, requiring a smaller degree of fusion among different classes of methods. In this section, we describe the architectures used to establish more realistic expectations on EQA performance and obtain initial results on VideoNavQA: several essential baselines and new models inspired by previous successes in visual question answering and computer vision.


\textbf{Language models} \enskip Question-only models have been surprisingly effective in EQA, often performing better than the complex approaches initially suggested~\cite{anand2018blindfold}. We include two simple yet powerful language modelling approaches in our evaluation: a \emph{1-layer LSTM}~\cite{hochreiter1997long} and an \emph{bag-of-words (BoW)}~\cite{ren2015exploring} model. Both of them contain an initial embedding layer where a separate representation is learned for each of the 134 vocabulary words. The BoW then averages all representations of the question words and feeds the result to a linear layer, while the LSTM encodes the entire sequence into a vector. These baselines demonstrate the inherent biases that exist in the environment distribution and place a lower bound on the desired performance of models that can usefully exploit visual information in this setting.

\textbf{Vision models} \enskip Learning to answer a question based only on the visual input is not expected to perform better than being biased towards the most frequently occurring answer (approximately 66\% of the VideoNavQA questions require a binary answer). To illustrate the expected behavior, we use a per-frame processing VGG-style~\cite{simonyan2014very} convolutional neural network with an LSTM classifier (V-CNN2D) and a C3D-like~\cite{tran2015learning} CNN (V-CNN3D).

\textbf{Concatenation models} \enskip Based on single-modality baseline results, we integrate the LSTM model with each of the two vision models, in order to derive a joint representation of the obtained features. We achieve this by concatenating the final question representation with their respective video representations (Concat-CNN2D/3D; see Figure~\ref{fig:concats}), then passing them through an MLP and obtaining the most likely answer via softmax.

\begin{figure}[t]
    \centering
    \includegraphics[trim={0 280 220 32},clip,width=0.5\linewidth]{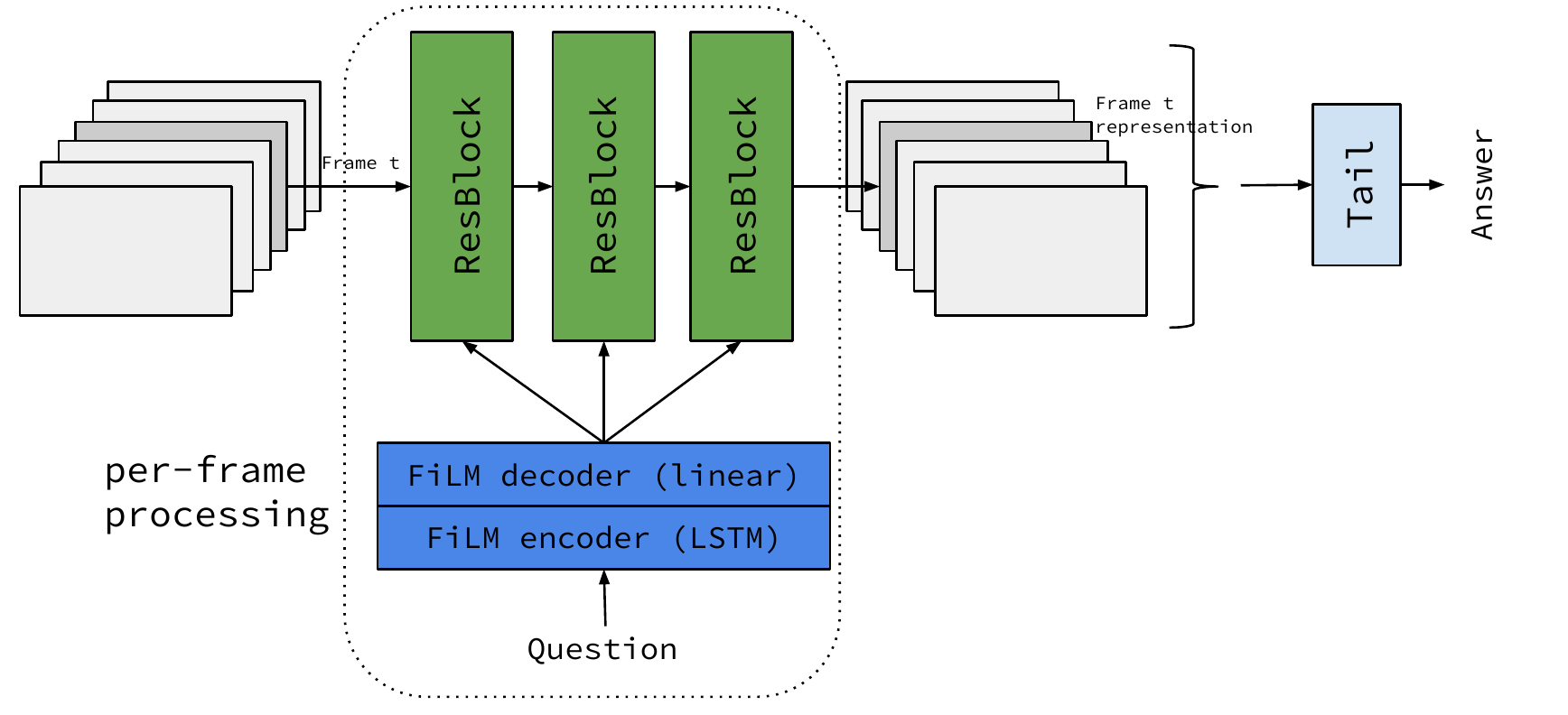}\includegraphics[trim={0 237 230 32},clip,width=0.5\linewidth]{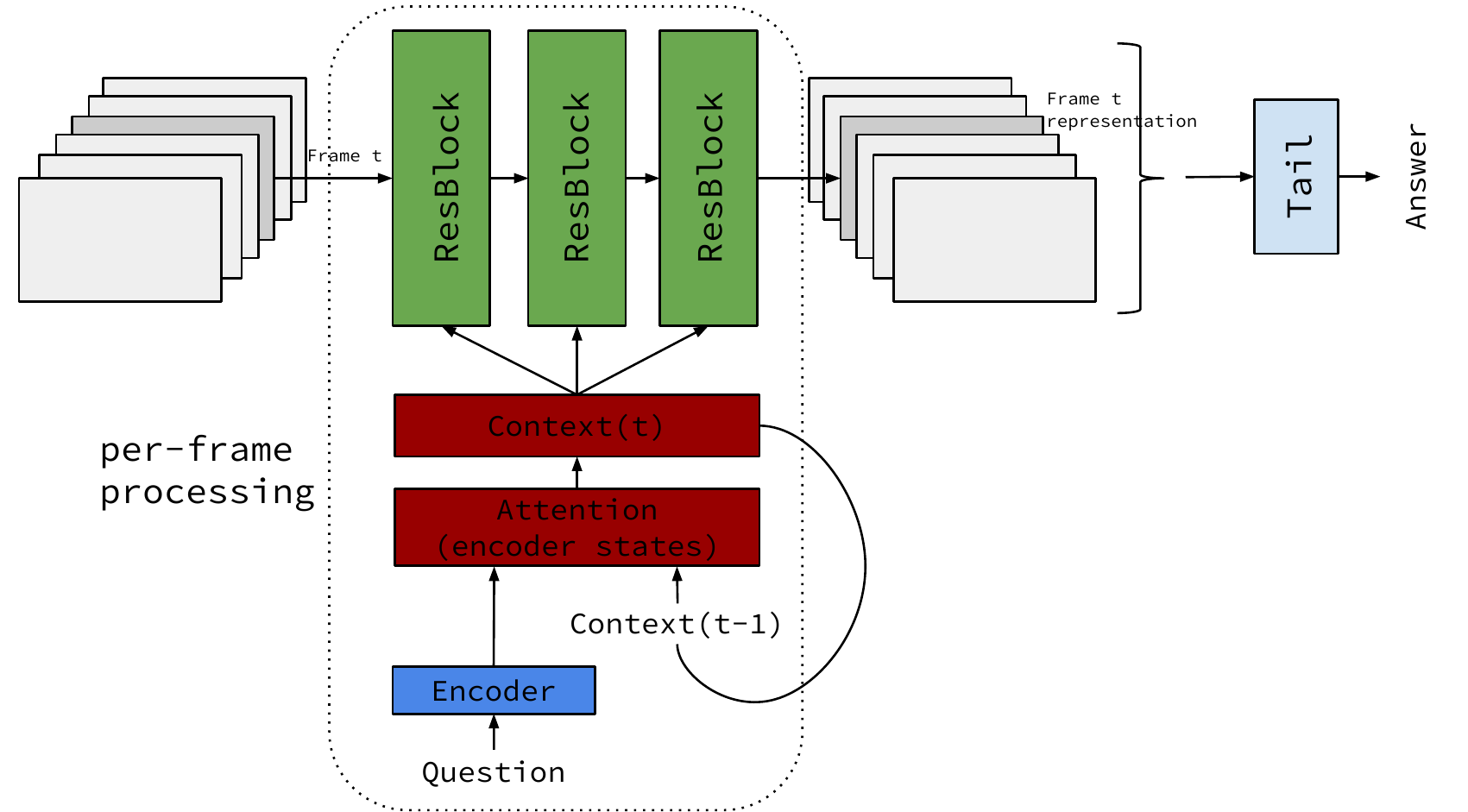}
    \caption{(Left:) Per-frame FiLM model. Video frames are processed separately by the ResBlocks, then all features are aggregated by the classifier to answer the question. (Right:) Temporal Multi-hop model. Each video frame is processed by the ResBlocks: the FiLM parameters are computed from the current attention context, which is initialized with the one from the previous frame. Temporal summarization is achieved via global max-pooling.}
    \vspace{-1em}
    \label{fig:films}
\end{figure}

\textbf{FiLM-based per-frame reasoning} \enskip Feature-wise linear modulation (FiLM)~\cite{perez2018film} has previously achieved state-of-the-art results on visual question answering tasks, directly using the question embedding to scale and shift the feature maps of a CNN pipeline taking the image as input. Here, we \emph{extend FiLM to address the additional VideoNavQA temporal dimension}. As per Figure~\ref{fig:films}, each frame is processed independently by a fixed number of ResBlocks~\cite{perez2018film}. FiLM conditioning ensures that visual information relevant for answering the question will be propagated to the final frame representation. We thereby obtain a series of visual features that we then aggregate via two mechanisms:
\begin{itemize}
    \item \emph{attention}: we apply a linear transformation to the features from each frame, then use recurrent attention~\cite{bahdanau2014neural, cho2014learning} to obtain an attended, final encoding. (FiLM AT)
    \item \emph{global max-pooling}: we apply a $1\times1$ convolution, followed by a feature-wise max-pooling operation over all frames that produces the final representation. (FiLM GP)
\end{itemize}
The motivation for these two classifiers is related to having a rich representation of the video across time, from which the visual reasoning method needs to select the information that is required to answer the question, typically contained in only a few frames. Attention and global max-pooling are generally effective at picking the essential features from a sequence and thus are reasonable choices for our model design.

\textbf{MAC} Compositional Attention Networks (MAC) \cite{hudson2018compositional} have achieved excellent performance in several VQA tasks \cite{hudson2018compositional,hudson2019gqa,bahdanau2018systematic}. Here we adapt this by applying a MAC model after a 2D-CNN and integrating it over time using an LSTM.

\textbf{Multi-hop temporal processing} \enskip The multi-hop extension of FiLM~\cite{strub2018visual} modulates feature maps at a particular level in the CNN hierarchy (i.e. the output of a certain ResBlock) by attending over the hidden states of the question encoder. The attention mechanism is initialized with the context vector computed for the previous level, which makes this approach scale better to settings with a longer input sequence such as a dialogue.

VideoNavQA requires the incorporation of an additional dimension in the visual reasoning process.  We propose a \emph{temporal multi-hop} model (see Figure~\ref{fig:films}) to condition the video features using the question input. We compute FiLM parameters for all ResBlocks at frame~$t$ by initializing the attention context with the one from frame $t-1$, attending over question encoder hidden states, obtaining the new context vector and passing it through a linear layer. This way, the FiLM parameters at a certain time step are based on what has already been computed for previous frames, thus modelling the temporal structure of the input.

\section{Experiments}

\subsection{Setup}

We evaluate all of the models described in the previous section on the VideoNavQA task and produce an initial expectation of the feasibility of EQA, using accuracy to compare the overall performance between different architectures. More in-depth analysis is then carried out for each of the models, in order to determine their respective strengths across different question categories and types. All models are trained with the Adam optimizer~\cite{kingma2014adam}, by monitoring the validation accuracy. Hyperparameters are reported after evaluating different combinations of layer hidden sizes, number of residual blocks and classifier dimensions.

\textbf{Language-only models} \enskip The recurrent model has an embedding layer of 512 units and an LSTM with 128 hidden units, whereas the BoW model uses an embedding size of 128. We use a batch size of 1024 and learning rates of $5e^{-5}$ and $1e^{-5}$, respectively.

\textbf{Video-only models} \enskip The V-CNN3D model has 3 \{convolutional, max-pool, batch normalization (BN)\} blocks with 64, 128, and 128 output feature maps, respectively, kernel size $(1,2,2)$ for the first block and $(4,4,4)$ otherwise. The classifier has 2 linear layers with 2048 and 128 hidden units respectively. BN~\cite{ioffe2015batch} and ReLU~\cite{glorot2011deep} activations are used for each layer. For the V-CNN2D model, we use a VGG configuration with 5 \{convolutional, max-pool\} blocks with 16, 32, 64, 128 and 128 channels, respectively, and kernel size 2.

\textbf{FiLM-extended models} \enskip The models that process each frame in a FiLM fashion have 4 (GP) or 5 (AT) ResBlocks with 1024 channels, preceded by a $1\times1$ convolution on the input. The attention mechanism for FiLM AT has 128 hidden units, whereas the global max-pooling classifier obtains 32 channels via the $1\times1$ convolution. The temporal multi-hop model has 3 ResBlocks and 64 tail channels. It is trained with a learning rate of $5e^{-5}$ and a batch size of 16, whereas the other two use $1e^{-4}$ and 32.

\textbf{Video dimensionality} \enskip The videos have an initial size of $140\times3\times160\times208$. In order to make learning feasible time-wise and allow the exploration of models with bigger capacity, we extract features from an object detector pre-trained on a set of 2000 frames that are not part of the dataset, which we initialize with the output from the 10$^{\text{th}}$ layer of a Faster R-CNN~\cite{ren2015faster}. The object detector has 3 \{conv, conv, BN, max-pool\} blocks with 512 output maps for each layer; the classifier consists of a 1024-unit linear layer followed by softmax. No significant difference in validation accuracy is noticed when training the Concat-CNN3D model with and without pre-trained features. We also adopt a randomized sub-sampling strategy to reduce the maximum video length from 140 to 35. On each training iteration, every section of 4 frames is reduced to 1 by randomly picking out a frame from the chunk---this enables models to become more robust and invariant to changes between nearby frames.

\subsection{Results}

\textbf{Model performance} \enskip As expected, the video-only baselines can only predict the most prevalent answer (\emph{Yes/True}). The LSTM proves to be the most powerful language model, achieving 7.5\% more than the bag-of-words. Surprisingly, Concat-CNN2D manages to outperform the other models, obtaining an accuracy of 64.47\%. The overall performance of Concat-CNN3D and FiLM AT is roughly similar, whereas FiLM GP achieves approximately 0.7\% less than Concat-CNN2D. The temporal multi-hop falls about 0.3\% behind FiLM GP, nevertheless outperforming the LSTM by 7\%.

Overall, these results suggest that the models presented manage to exploit the visual context of the environment, improving by a significant margin over the language-only capabilities of the LSTM and BoW. This finding represents an initial validation of the feasibility of the VideoNavQA task, ensuring that the manner in which we constructed the dataset allows the visual reasoning process to be generalizable across new environments. A more focused investigation of novel VQA techniques is further required to accommodate the rich dimensionality and contextual information encountered in the videos.

\begin{table}
\begin{center}
\begin{tabular}{|c|c|ccc|}
\hline
 \small Model & \small Accuracy All & \small Yes/No & \small Other & \small Num \\
\hline

\small	BoW	  &	\small	49.02	&	\small	57.67	&	\small	30.57	&	\small 40.21	 \\
\small	LSTM	&	\small	56.49	&	\small	68.36	&	\small	35.27	&	\small 38.90	\\\hline
\small	V-CNN3D	&	\small	33.29	&	\small	-	&	\small	-	&	\small -	\\
\small	V-CNN2D	&	\small	33.62	&	\small	-	&	\small	-	&	\small	-\\\hline				
\small	Concat-CNN3D	&	\small	64.00	&	\small	72.99	&	\small	49.12	&	\small 49.10	\\
\small	Concat-CNN2D	&	\small	64.47	&	\small	73.50	&	\small	49.20	&	\small 49.59	\\
\small	FiLM-GP	&	\small	63.79	&	\small	72.91	&	\small	47.71	&	\small 50.00	\\
\small	FiLM-AT	&	\small	64.08	&	\small	72.93	&	\small	49.54	&	\small 49.26	\\
\small	Temporal multi-hop	&	\small	63.53	&	\small	71.81	&	\small 49.54		&	\small 50.16	\\
\small	MAC	&	\small	62.32	&	\small	69.02	&	\small	51.37	&	\small	50.99 \\
\hline
\end{tabular}
\end{center}
\vspace{1em}
\caption{Results for all models reported in terms of accuracy on the VideoNavQA test set. We also use standard VQA reporting of Yes/No, other, and number categories~\cite{antol2015vqa}.}
\label{table:results}
\end{table}

\begin{figure}[t]
    \centering
    \includegraphics[width=0.25\linewidth]{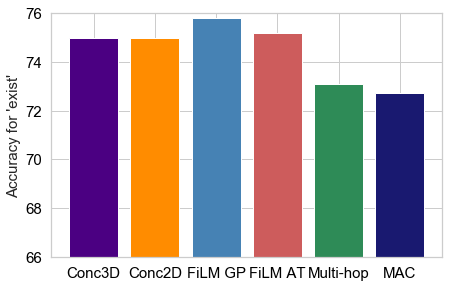}\includegraphics[width=0.25\linewidth]{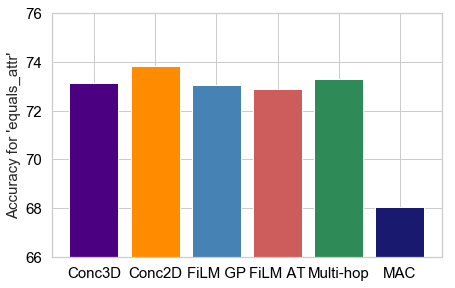}\includegraphics[width=0.25\linewidth]{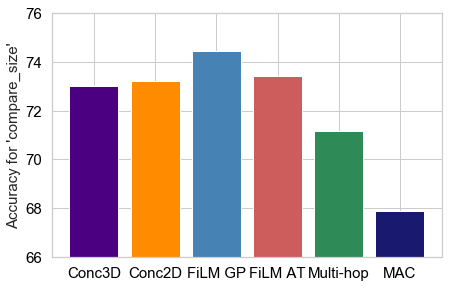}\includegraphics[width=0.25\linewidth]{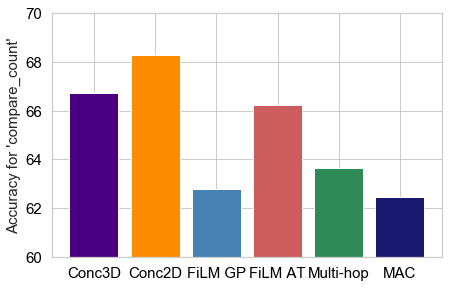}\\
    \centering
    \includegraphics[width=0.25\linewidth]{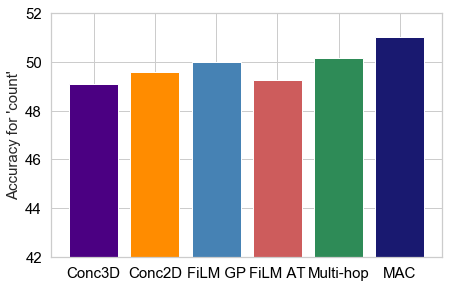}\includegraphics[width=0.25\linewidth]{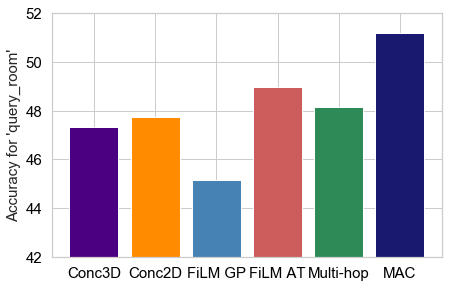}\includegraphics[width=0.25\linewidth]{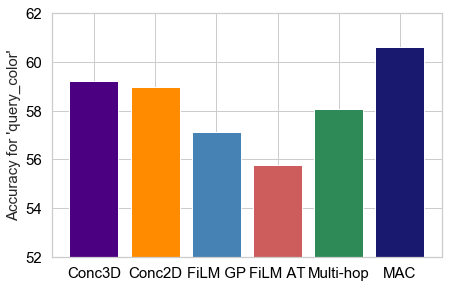}\includegraphics[width=0.25\linewidth]{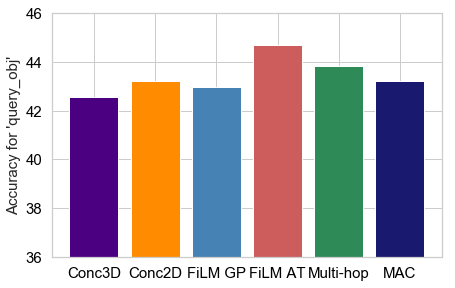}
    \caption{Comparative performance of the models on each question category. Questions from the top row categories have binary answers, whereas the ones from the bottom row are answered by integer counts, room types, colors and object types, respectively.}
    \label{fig:accsperqcat}
\end{figure}

\textbf{Analysis by question category} \enskip Figure~\ref{fig:accsperqcat} provides a more detailed analysis of the model performances on the test set across the eight question categories, in order to reveal the capabilities developed on VideoNavQA sub-tasks. All models score highest on \emph{existence} questions, with FiLM GP having an edge over others, whereas \emph{identifying object types} seems to be the hardest task: Concat-CNN3D performs worst, whereas FiLM AT achieves the highest accuracy, closely followed by Multi-hop. MAC is by far the best at \emph{identifying the room locations of objects} and obtains the strongest performance on \emph{counting} as well. Here, all other models score roughly 50\%. When \emph{comparing object attributes} (i.e. room location, color), most models achieve around 73\%, with Concat-CNN2D obtaining a 1\% edge and MAC, a 5\% drop. MAC surpasses all other models when \emph{identifying colors}, with Concat-CNNs achieving the next best result of 58\% and FiLM AT performing worst, just under 56\%. FiLM GP does relatively well at \emph{comparing object/room sizes}, followed by FiLM AT, whereas Multi-Hop obtains a weaker performance of around 71\% and MAC achieves only 68\%. Alternatively, FiLM GP and MAC do much worse than all the other models when \emph{comparing counts} and Concat-CNN2D considerably outperforms them, with an overall performance gap of 5\%.

\section{Conclusion}

We have introduced the VideoNavQA task and dataset as a means to gain a better understanding of what is achievable in the EQA domain, where the environment context is visually rich---here, suitable methods need to be integrated accordingly to build an agent that can reason, navigate and act. This task represents an alternative view of the EQA paradigm: the navigation aspect is made trivial by providing nearly-optimal trajectories to the agent, but the reasoning difficulty becomes much higher, having 28 question types across 8 categories.

By reporting initial results from a variety of single-modality and integrative models, we establish a lower bound on the achievable performance in the VideoNavQA task. Since results are not widely different across architectures, we note the need for more investigation into whether there exist other video QA models that can better isolate the required visual information, bridging the gap to human performance. By first tackling the VideoNavQA task effectively, the research community will allow more clear and measured progress when attempting to integrate navigation into the EQA task, gradually reaching a better solution.

\section*{Acknowledgements}

We wish to thank Ankesh Anand and Ethan Perez for the useful discussions over the course of this project. CC is funded by DREAM CDT and was supported by Mila during the time in Montr\'{e}al. EB is funded by IVADO. We also thank the University of Cambridge Research Computing Services for providing HPC cluster resources.
%

\bibliography{egbib}

\begin{thebibliography}{40}
\providecommand{\natexlab}[1]{#1}
\providecommand{\url}[1]{\texttt{#1}}
\expandafter\ifx\csname urlstyle\endcsname\relax
  \providecommand{\doi}[1]{doi: #1}\else
  \providecommand{\doi}{doi: \begingroup \urlstyle{rm}\Url}\fi

\bibitem[Anand et~al.(2018)Anand, Belilovsky, Kastner, Larochelle, and
  Courville]{anand2018blindfold}
Ankesh Anand, Eugene Belilovsky, Kyle Kastner, Hugo Larochelle, and Aaron
  Courville.
\newblock Blindfold {B}aselines for {E}mbodied {QA}.
\newblock \emph{arXiv preprint arXiv:1811.05013}, 2018.

\bibitem[Antol et~al.(2015)Antol, Agrawal, Lu, Mitchell, Batra,
  Lawrence~Zitnick, and Parikh]{antol2015vqa}
Stanislaw Antol, Aishwarya Agrawal, Jiasen Lu, Margaret Mitchell, Dhruv Batra,
  C~Lawrence~Zitnick, and Devi Parikh.
\newblock {VQA}: Visual question answering.
\newblock In \emph{Proceedings of the IEEE international conference on computer
  vision}, pages 2425--2433, 2015.

\bibitem[Bahdanau et~al.(2014)Bahdanau, Cho, and Bengio]{bahdanau2014neural}
Dzmitry Bahdanau, Kyunghyun Cho, and Yoshua Bengio.
\newblock Neural machine translation by jointly learning to align and
  translate.
\newblock \emph{arXiv preprint arXiv:1409.0473}, 2014.

\bibitem[Bahdanau et~al.(2018)Bahdanau, Murty, Noukhovitch, Nguyen, de~Vries,
  and Courville]{bahdanau2018systematic}
Dzmitry Bahdanau, Shikhar Murty, Michael Noukhovitch, Thien~Huu Nguyen, Harm
  de~Vries, and Aaron Courville.
\newblock Systematic {G}eneralization: {W}hat {I}s {R}equired and {C}an {I}t
  {B}e {L}earned?
\newblock \emph{arXiv preprint arXiv:1811.12889}, 2018.

\bibitem[Cadene et~al.(2019)Cadene, Ben-Younes, Thome, and
  Cord]{Cadene_2019_CVPR}
Remi Cadene, Hedi Ben-Younes, Nicolas Thome, and Matthieu Cord.
\newblock {MUREL}: {M}ultimodal {R}elational {R}easoning for {V}isual
  {Q}uestion {A}nswering.
\newblock In \emph{{IEEE} Conference on Computer Vision and Pattern Recognition
  {CVPR}}, 2019.
\newblock URL \url{http://remicadene.com/pdfs/paper_cvpr2019.pdf}.

\bibitem[Cho et~al.(2014)Cho, Van~Merri{\"e}nboer, Gulcehre, Bahdanau,
  Bougares, Schwenk, and Bengio]{cho2014learning}
Kyunghyun Cho, Bart Van~Merri{\"e}nboer, Caglar Gulcehre, Dzmitry Bahdanau,
  Fethi Bougares, Holger Schwenk, and Yoshua Bengio.
\newblock Learning phrase representations using {RNN} encoder-decoder for
  statistical machine translation.
\newblock \emph{arXiv preprint arXiv:1406.1078}, 2014.

\bibitem[Das et~al.(2018{\natexlab{a}})Das, Datta, Gkioxari, Lee, Parikh, and
  Batra]{embodiedqa}
Abhishek Das, Samyak Datta, Georgia Gkioxari, Stefan Lee, Devi Parikh, and
  Dhruv Batra.
\newblock {{E}mbodied {Q}uestion {A}nswering}.
\newblock In \emph{Proceedings of the IEEE Conference on Computer Vision and
  Pattern Recognition (CVPR)}, 2018{\natexlab{a}}.

\bibitem[Das et~al.(2018{\natexlab{b}})Das, Gkioxari, Lee, Parikh, and
  Batra]{das2018neural}
Abhishek Das, Georgia Gkioxari, Stefan Lee, Devi Parikh, and Dhruv Batra.
\newblock {Neural Modular Control for Embodied Question Answering}.
\newblock \emph{arXiv preprint arXiv:1810.11181}, 2018{\natexlab{b}}.

\bibitem[Fukui et~al.(2016)Fukui, Park, Yang, Rohrbach, Darrell, and
  Rohrbach]{fukui2016multimodal}
Akira Fukui, Dong~Huk Park, Daylen Yang, Anna Rohrbach, Trevor Darrell, and
  Marcus Rohrbach.
\newblock Multimodal compact bilinear pooling for visual question answering and
  visual grounding.
\newblock \emph{arXiv preprint arXiv:1606.01847}, 2016.

\bibitem[Glorot et~al.(2011)Glorot, Bordes, and Bengio]{glorot2011deep}
Xavier Glorot, Antoine Bordes, and Yoshua Bengio.
\newblock Deep sparse rectifier neural networks.
\newblock In \emph{Proceedings of the fourteenth international conference on
  artificial intelligence and statistics}, pages 315--323, 2011.

\bibitem[Gordon et~al.(2018)Gordon, Kembhavi, Rastegari, Redmon, Fox, and
  Farhadi]{gordon2018iqa}
Daniel Gordon, Aniruddha Kembhavi, Mohammad Rastegari, Joseph Redmon, Dieter
  Fox, and Ali Farhadi.
\newblock {IQA: Visual question answering in interactive environments}.
\newblock In \emph{Proceedings of the IEEE Conference on Computer Vision and
  Pattern Recognition}, pages 4089--4098, 2018.

\bibitem[Hochreiter and Schmidhuber(1997)]{hochreiter1997long}
Sepp Hochreiter and J{\"u}rgen Schmidhuber.
\newblock Long short-term memory.
\newblock \emph{Neural computation}, 9\penalty0 (8):\penalty0 1735--1780, 1997.

\bibitem[Hu et~al.(2017)Hu, Andreas, Rohrbach, Darrell, and
  Saenko]{hu2017learning}
Ronghang Hu, Jacob Andreas, Marcus Rohrbach, Trevor Darrell, and Kate Saenko.
\newblock {Learning to reason: End-to-end module networks for visual question
  answering}.
\newblock In \emph{Proceedings of the IEEE International Conference on Computer
  Vision}, pages 804--813, 2017.

\bibitem[Hudson and Manning(2018)]{hudson2018compositional}
Drew~A Hudson and Christopher~D Manning.
\newblock Compositional {A}ttention {N}etworks for {M}achine {R}easoning.
\newblock \emph{arXiv preprint arXiv:1803.03067}, 2018.

\bibitem[Hudson and Manning(2019)]{hudson2019gqa}
Drew~A Hudson and Christopher~D Manning.
\newblock {GQA}: a new dataset for compositional question answering over
  real-world images.
\newblock \emph{arXiv preprint arXiv:1902.09506}, 2019.

\bibitem[Ioffe and Szegedy(2015)]{ioffe2015batch}
Sergey Ioffe and Christian Szegedy.
\newblock Batch normalization: Accelerating deep network training by reducing
  internal covariate shift.
\newblock \emph{arXiv preprint arXiv:1502.03167}, 2015.

\bibitem[Johnson et~al.(2017)Johnson, Hariharan, van~der Maaten, Fei-Fei,
  Lawrence~Zitnick, and Girshick]{johnson2017clevr}
Justin Johnson, Bharath Hariharan, Laurens van~der Maaten, Li~Fei-Fei,
  C~Lawrence~Zitnick, and Ross Girshick.
\newblock {CLEVR}: A diagnostic dataset for compositional language and
  elementary visual reasoning.
\newblock In \emph{Proceedings of the IEEE Conference on Computer Vision and
  Pattern Recognition}, pages 2901--2910, 2017.

\bibitem[Kingma and Ba(2014)]{kingma2014adam}
Diederik~P Kingma and Jimmy Ba.
\newblock {Adam: A method for stochastic optimization}.
\newblock \emph{arXiv preprint arXiv:1412.6980}, 2014.

\bibitem[Kolve et~al.(2017)Kolve, Mottaghi, Han, VanderBilt, Weihs, Herrasti,
  Gordon, Zhu, Gupta, and Farhadi]{ai2thor}
Eric Kolve, Roozbeh Mottaghi, Winson Han, Eli VanderBilt, Luca Weihs, Alvaro
  Herrasti, Daniel Gordon, Yuke Zhu, Abhinav Gupta, and Ali Farhadi.
\newblock {AI2-THOR: An Interactive 3D Environment for Visual AI}.
\newblock \emph{arXiv}, 2017.

\bibitem[Lei et~al.(2018)Lei, Yu, Bansal, and Berg]{lei2018tvqa}
Jie Lei, Licheng Yu, Mohit Bansal, and Tamara~L Berg.
\newblock {TVQA: Localized, Compositional Video Question Answering}.
\newblock In \emph{EMNLP}, 2018.

\bibitem[Malinowski et~al.(2018)Malinowski, Doersch, Santoro, and
  Battaglia]{malinowski2018learning}
Mateusz Malinowski, Carl Doersch, Adam Santoro, and Peter Battaglia.
\newblock Learning visual question answering by bootstrapping hard attention.
\newblock In \emph{Proceedings of the European Conference on Computer Vision
  (ECCV)}, pages 3--20, 2018.

\bibitem[Mishkin et~al.(2019)Mishkin, Dosovitskiy, and
  Koltun]{mishkin2019benchmarking}
Dmytro Mishkin, Alexey Dosovitskiy, and Vladlen Koltun.
\newblock {Benchmarking Classic and Learned Navigation in Complex 3D
  Environments}.
\newblock \emph{arXiv preprint arXiv:1901.10915}, 2019.

\bibitem[Mun et~al.(2017)Mun, Seo, Jung, and Han]{mun2017marioQA}
Jonghwan Mun, Paul~Hongsuck Seo, Ilchae Jung, and Bohyung Han.
\newblock {MarioQA: Answering Questions by Watching Gameplay Videos}.
\newblock In \emph{ICCV}, 2017.

\bibitem[Narasimhan et~al.(2018)Narasimhan, Lazebnik, and
  Schwing]{narasimhan2018out}
Medhini Narasimhan, Svetlana Lazebnik, and Alexander Schwing.
\newblock Out of the box: {R}easoning with graph convolution nets for factual
  visual question answering.
\newblock In \emph{Advances in Neural Information Processing Systems}, pages
  2654--2665, 2018.

\bibitem[Nguyen et~al.(2018)Nguyen, Sharma, Schulz, and Asri]{nguyen2018film}
Dat~Tien Nguyen, Shikhar Sharma, Hannes Schulz, and Layla~El Asri.
\newblock From {F}i{LM} to {V}ideo: {M}ulti-turn {Q}uestion {A}nswering with
  {M}ulti-modal {C}ontext.
\newblock \emph{arXiv preprint arXiv:1812.07023}, 2018.

\bibitem[Norcliffe-Brown et~al.(2018)Norcliffe-Brown, Vafeias, and
  Parisot]{norcliffe2018learning}
Will Norcliffe-Brown, Stathis Vafeias, and Sarah Parisot.
\newblock Learning conditioned graph structures for interpretable visual
  question answering.
\newblock In \emph{Advances in Neural Information Processing Systems}, pages
  8334--8343, 2018.

\bibitem[Perez et~al.(2018)Perez, Strub, De~Vries, Dumoulin, and
  Courville]{perez2018film}
Ethan Perez, Florian Strub, Harm De~Vries, Vincent Dumoulin, and Aaron
  Courville.
\newblock {FiLM}: {V}isual reasoning with a general conditioning layer.
\newblock In \emph{Thirty-Second AAAI Conference on Artificial Intelligence},
  2018.

\bibitem[Ren et~al.(2015{\natexlab{a}})Ren, Kiros, and Zemel]{ren2015exploring}
Mengye Ren, Ryan Kiros, and Richard Zemel.
\newblock Exploring models and data for image question answering.
\newblock In \emph{Advances in neural information processing systems}, pages
  2953--2961, 2015{\natexlab{a}}.

\bibitem[Ren et~al.(2015{\natexlab{b}})Ren, He, Girshick, and
  Sun]{ren2015faster}
Shaoqing Ren, Kaiming He, Ross Girshick, and Jian Sun.
\newblock {Faster R-CNN: Towards real-time object detection with region
  proposal networks}.
\newblock In \emph{Advances in neural information processing systems}, pages
  91--99, 2015{\natexlab{b}}.

\bibitem[Savva et~al.(2019)Savva, Kadian, Maksymets, Zhao, Wijmans, Jain,
  Straub, Liu, Koltun, Malik, et~al.]{savva2019habitat}
Manolis Savva, Abhishek Kadian, Oleksandr Maksymets, Yili Zhao, Erik Wijmans,
  Bhavana Jain, Julian Straub, Jia Liu, Vladlen Koltun, Jitendra Malik, et~al.
\newblock {Habitat: A Platform for Embodied AI Research}.
\newblock \emph{arXiv preprint arXiv:1904.01201}, 2019.

\bibitem[Simonyan and Zisserman(2014)]{simonyan2014very}
Karen Simonyan and Andrew Zisserman.
\newblock Very deep convolutional networks for large-scale image recognition.
\newblock \emph{arXiv preprint arXiv:1409.1556}, 2014.

\bibitem[Song et~al.(2017)Song, Yu, Zeng, Chang, Savva, and
  Funkhouser]{song2016ssc}
Shuran Song, Fisher Yu, Andy Zeng, Angel~X Chang, Manolis Savva, and Thomas
  Funkhouser.
\newblock {Semantic Scene Completion from a Single Depth Image}.
\newblock \emph{Proceedings of 30th IEEE Conference on Computer Vision and
  Pattern Recognition}, 2017.

\bibitem[Strub et~al.(2018)Strub, Seurin, Perez, De~Vries, Mary, Preux,
  Courville, and Pietquin]{strub2018visual}
Florian Strub, Mathieu Seurin, Ethan Perez, Harm De~Vries, J{\'e}r{\'e}mie
  Mary, Philippe Preux, Aaron Courville, and Olivier Pietquin.
\newblock Visual {R}easoning with {M}ulti-hop {F}eature {M}odulation.
\newblock In \emph{Proceedings of the European Conference on Computer Vision
  (ECCV)}, pages 784--800, 2018.

\bibitem[Tapaswi et~al.(2016)Tapaswi, Zhu, Stiefelhagen, Torralba, Urtasun, and
  Fidler]{MovieQA}
Makarand Tapaswi, Yukun Zhu, Rainer Stiefelhagen, Antonio Torralba, Raquel
  Urtasun, and Sanja Fidler.
\newblock {MovieQA: Understanding Stories in Movies through
  Question-Answering}.
\newblock In \emph{IEEE Conference on Computer Vision and Pattern Recognition
  (CVPR)}, 2016.

\bibitem[Teney et~al.(2017)Teney, Liu, and van~den Hengel]{teney2017graph}
Damien Teney, Lingqiao Liu, and Anton van~den Hengel.
\newblock Graph-structured representations for visual question answering.
\newblock In \emph{Proceedings of the IEEE Conference on Computer Vision and
  Pattern Recognition}, pages 1--9, 2017.

\bibitem[Tran et~al.(2015)Tran, Bourdev, Fergus, Torresani, and
  Paluri]{tran2015learning}
Du~Tran, Lubomir Bourdev, Rob Fergus, Lorenzo Torresani, and Manohar Paluri.
\newblock Learning spatiotemporal features with {3D} convolutional networks.
\newblock In \emph{Proceedings of the IEEE international conference on computer
  vision}, pages 4489--4497, 2015.

\bibitem[Wijmans et~al.(2019)Wijmans, Datta, Maksymets, Das, Gkioxari, Lee,
  Essa, Parikh, and Batra]{eqa_matterport}
Erik Wijmans, Samyak Datta, Oleksandr Maksymets, Abhishek Das, Georgia
  Gkioxari, Stefan Lee, Irfan Essa, Devi Parikh, and Dhruv Batra.
\newblock {E}mbodied {Q}uestion {A}nswering in {P}hotorealistic {E}nvironments
  with {P}oint {C}loud {P}erception.
\newblock In \emph{Proceedings of the IEEE Conference on Computer Vision and
  Pattern Recognition (CVPR)}, 2019.

\bibitem[Wu et~al.(2018)Wu, Wu, Gkioxari, and Tian]{wu2018building}
Yi~Wu, Yuxin Wu, Georgia Gkioxari, and Yuandong Tian.
\newblock {Building generalizable agents with a realistic and rich 3D
  environment}.
\newblock \emph{arXiv preprint arXiv:1801.02209}, 2018.

\bibitem[Yang et~al.(2016)Yang, He, Gao, Deng, and Smola]{yang2016stacked}
Zichao Yang, Xiaodong He, Jianfeng Gao, Li~Deng, and Alex Smola.
\newblock Stacked attention networks for image question answering.
\newblock In \emph{Proceedings of the IEEE conference on computer vision and
  pattern recognition}, pages 21--29, 2016.

\bibitem[Yu et~al.(2019)Yu, Chen, Gkioxari, Bansal, Berg, and
  Batra]{eqa_multitarget}
Licheng Yu, Xinlei Chen, Georgia Gkioxari, Mohit Bansal, Tamara~L. Berg, and
  Dhruv Batra.
\newblock {Multi-Target Embodied Question Answering}.
\newblock In \emph{Proceedings of the IEEE Conference on Computer Vision and
  Pattern Recognition (CVPR)}, 2019.

\end{thebibliography}
\newpage

\section*{Appendix}

\subsection*{Detailed breakdown of question templates and respective counts}

\begin{itemize}
    \item \textbf{Equals<attr>}\\ 
        'Are all <attr> <obj\_type-pl> <color>?': 4014\\
        'Are all <attr> <obj\_type-pl> in the <room\_type>?': 3811\\
        'Are all <attr> things <obj\_type-pl>?': 3539\\
        'Are both the <attr1> <obj\_type1> and the <attr2> <obj\_type2> <color>?': 3968\\
        'Are both the <attr1> <obj\_type1> and the <attr2> <obj\_type2> in the <room\_type>?': 3804\\
        'Are the <attr1> <obj\_type1> and the <attr2> <obj\_type2> the same color?': 4018\\
        'Is the <attr1> thing <rel> the <attr2> <obj\_type2> <art> <obj\_type1>?': 3315
    \item \textbf{Count}\\ 
         'How many <attr1> <obj\_type1-pl> are in the room containing the <attr2> <obj\_type2>?': 3999\\
         'How many <attr> <obj\_type-pl> are in the <room\_type>?': 3763\\
         'How many <obj\_type-pl> are <attr>?': 4120\\
         'How many rooms have <attr> <obj\_type-pl>?': 3834
    \item \textbf{Compare<count>}\\ 
         'Are there <comp> <attr1> <obj\_type1-pl> than <attr2> <obj\_type2-pl>?': 4058\\
         'Are there as many <attr1> <obj\_type1-pl> as there are <attr2> <obj\_type2-pl>?': 4100
    \item \textbf{Compare<size>}\\ 
         'Is the <attr1> <obj\_type> <comp\_rel> than the <attr2> one?': 3272\\
         'Is the <room\_type1> <comp\_rel> than the <room\_type2>?': 3148
    \item \textbf{Exist}\\ 
         'Is there <art> <attr> <obj\_type>?': 4122\\
         'Is there <art> <room\_type>?': 3335\\
         'Is there a room that has set(<art> <attr\{\}> <obj\_type\{\}>)?': 3877\\
         'Is there set(<art> <attr\{\}> <obj\_type\{\}>) in the <room\_type>?': 4025\\
         'Is there set(<art> <attr\{\}> <obj\_type\{\}>)?': 4107\\
         'Is there set(<art> <room\_type\{\}>)?': 3750
    \item \textbf{Query<color>}\\ 
         'What color is the <attr1> <obj\_type1> <rel> the <attr2> <obj\_type2>?': 2178\\
         'What color is the <attr> <obj\_type>?': 3592
    \item \textbf{Query<obj\_type>}\\ 
         'What is the <attr1> thing <rel> the <attr2> <obj\_type2>?': 3119\\
         'What is the <attr> thing?': 2883
    \item \textbf{Query<room\_location>}\\ 
         'Where are the set(<attr\{\}> <obj\_type\{\}>)?': 3816\\
         'Where is the <attr1> <obj\_type1> <rel> the <attr2> <obj\_type2>?': 2284\\
         'Where is the <attr> <obj\_type>?': 3481
\end{itemize}

\subsection*{Model hyperparameters}

In order to find the best performance on VideoNavQA, we have ran several combinations of hyperparameters for each of the described models. We detail settings evaluated on the validation set below. The average running time per epoch for the visual reasoning models is approximately 5 hours on a 16GB Tesla P100 GPU.

\textbf{LSTM} \enskip Embedding size: \{128, 256, 512, 1024\}. Learning rate: \{$5e^{-5}$, $1e^{-4}$\}.

\textbf{BoW} \enskip Embedding size: \{128, 256, 512\}. Learning rate: \{$1e^{-5}$, $5e^{-5}$\}.

\textbf{FiLM GP} \enskip Number of ResBlocks: \{3, 4, 5\}. Learning rate: \{$1e^{-4}$, $1e^{-3}$\}. Number of classifier channels: \{32, 64\}.

\textbf{FiLM AT} \enskip Number of ResBlocks: \{3, 4, 5\}. Attention hidden size: \{128, 256\}. Learning rate: \{$1e^{-5}$, $1e^{-4}$\}.

\textbf{Multi-hop} \enskip Number of ResBlocks: \{3, 4, 5\}. Learning rate: \{$1e^{-5}$, $1e^{-4}$\}. Number of classifier channels: \{32, 64\}.

\textbf{MAC} \enskip Number of CNN Layers: \{2,3\}. Width: \{512, 1024\}. MAC time steps: \{5,6\}. We adapt the ramp-up/down Adam learning schedule popularly used in VQA~\cite{Cadene_2019_CVPR}, ramping up the learning rate to $1e^{-4}$ in the first 2 epochs and then decaying it to $1e^{-5}$ after epoch 10 (training is done for a total of 15 epochs).

\end{document}